\documentclass{article}
\usepackage{spconf,amsmath,graphicx,hyperref}
\usepackage{amssymb,bm,booktabs,xcolor}

\title{STRUCTURED VISUAL TARGET LEARNING FOR\\
CROSS-SUBJECT EEG-TO-IMAGE RETRIEVAL}

\name{%
Salini Yadav$^{1}$\thanks{Corresponding author. Email:
\texttt{salini\_y1@cs.iitr.ac.in}},
Taveena Lotey$^{1}$,
Mickaël Coustaty$^{2}$,
Pravendra Singh$^{1}$,
Partha Pratim Roy$^{3}$
}

\address{%
$^{1}$Indian Institute of Technology Roorkee, India\\
$^{2}$University of La Rochelle, France\\
$^{3}$Indian Institute of Technology (ISM) Dhanbad, India
}

\begin{document}
\ninept
\maketitle

\begin{abstract}
Cross-subject EEG-to-image retrieval requires a neural representation trained on source subjects to remain aligned with a visual embedding space for an unseen subject. Whereas existing methods primarily focus on the EEG side, we address this problem from the perspective of the visual target. Our approach preserves the spatial information of the Perception Encoder, converts its patch grid into a compact set of learned visual views, and aggregates them for each image with a block-structured, content-dependent router. The target is learned jointly with the EEG encoder through contrastive learning with MMD regularization across source subjects. For deployment, we propose a training-free representation refinement that aligns frozen embeddings without updating either encoder. Under leave-one-subject-out evaluation on THINGS-EEG2, the structured target achieves 35.3\%/65.6\% Top-1/Top-5 accuracy, the best among compared methods. Refinement raises this to 48.1\%/77.1\%, an 18.5\% Top-1 gain over the strongest compared method, improving all ten held-out subjects
\footnote{Code: \url{https://github.com/Shalini-Y1/SVTL_EEG2Image}}.
\end{abstract}

% \begin{abstract}
% \textcolor{red}{Reduce abstract to 100–150 words. Cross-subject EEG-to-image retrieval requires a source-trained neural representation to remain compatible with a visual embedding space for an unseen subject.} We study this problem from the perspective of the visual target rather than introducing another subject-specific EEG model. Our approach preserves the spatial information of the Perception Encoder, converts its patch grid into a compact set of learned visual views, and combines these views using a block-structured content-dependent router. The resulting target is trained jointly with the EEG representation under a strict leave-one-subject-out protocol. We further evaluate a frozen, label-free deployment calibration stage that operates after source training and does not update either encoder. This calibration is included as a controlled deployment component rather than claimed as a new alignment algorithm. On THINGS-EEG2, the proposed structured visual target improves the ten-fold cross-subject baseline from 35.0\%/64.7\% to 35.3\%/65.6\% Top-1/Top-5. Adding the label-free deployment calibration raises performance to 48.1\%/77.1\%. The gain is observed for all ten held-out subjects. 
% \textcolor{red}{include attres along with PE}
% \end{abstract}


\section{Introduction}
\label{sec:intro}

Visual decoding from EEG aims to recover perceived visual content from non-invasive neural recordings \cite{gifford2022large,li2024visual}. EEG-to-image retrieval systems map neural responses and candidate images into a shared embedding space, enabling zero-shot recognition of concepts never seen as output classes during training \cite{li2024visual,song2024decoding}. On the THINGS-EEG2 benchmark \cite{gifford2022large}, progress depends on both the EEG encoder and the visual target it is trained to match \cite{li2024visual,jiang2026subject,wang2025neuroclip}.

A practically important but considerably harder setting is cross-subject retrieval. Under leave-one-subject-out (LOSO) evaluation, the model is trained on several subjects and tested on one whose recordings are never observed \cite{li2026cross}. Because EEG responses to the same image vary substantially across individuals, a mapping learned from source subjects often transfers poorly to a new one. Differences in head anatomy, electrode placement, and individual neural response patterns shift the EEG feature distribution, so an encoder that performs well on source subjects may produce embeddings that no longer align with the visual space for the target subject \cite{saha2020intra}. Closing this gap is essential for deployment, since subject-specific labeled calibration is costly in practice.

Prior work has addressed this problem mainly from the neural side. NICE established contrastive EEG-to-image retrieval on THINGS-EEG2 with frozen visual features \cite{song2024decoding}, and later methods improved the EEG encoder, multimodal supervision, or the choice of visual features \cite{li2024visual,tang2026aligning,wang2025neuroclip,11094846,zhang2026neurobridge,du2026deep}. SAMGA introduced subject-aware multi-granularity alignment for zero-shot EEG-to-image retrieval \cite{jiang2026subject}. Classical cross-subject BCI methods, such as Euclidean alignment, hyperalignment, and Riemannian Procrustes analysis, instead align the neural data directly \cite{he2019transfer,haxby2011common,rodrigues2018riemannian}, while test-time calibration adapts frozen models using unlabeled target-subject data \cite{huang2026sattc}. In most of these approaches, however, the visual side is represented by a single global image embedding shared by all subjects.

We argue that this visual target is itself a bottleneck for cross-subject generalization. A global embedding compresses an image into one vector, so every subject's EEG must be mapped onto the same summary, even though different individuals' responses may reflect different aspects of the image. Modern vision encoders retain far richer information: the Perception Encoder (PE) shows that intermediate spatial representations carry distributed, complementary visual information not confined to the final output \cite{bolya2026perception}. We hypothesize that supervising EEG with several complementary image views provides the encoder with multiple ways to match a new subject's response, thereby enabling better transfer across individuals.

\begin{figure*}[ht]
    \centering
    \includegraphics[width=\textwidth]{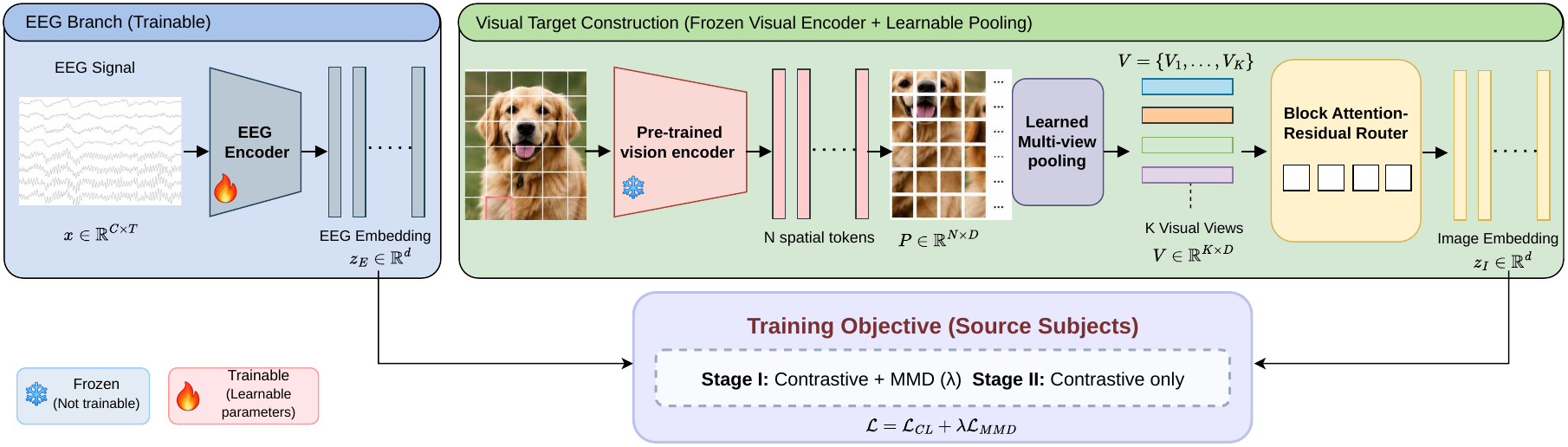}
    \caption{Overview of the proposed cross-subject EEG-to-image retrieval framework. The trainable EEG encoder maps EEG signals into the shared embedding space, while a frozen pre-trained vision encoder produces spatial visual tokens that are converted into multiple visual views using learned multi-view pooling and fused by the block attention--residual router. EEG and image embeddings are aligned through a two-stage objective, combining contrastive learning with MMD-based distribution alignment in Stage I and contrastive learning alone in Stage II.}
    \label{fig:architecture}
    \vspace{-0.5cm}
\end{figure*}

We therefore introduce a \emph{structured multi-view visual target} constructed from PE spatial features. Learned queries \cite{li2023blip2,jaegle2021perceiver} pool the spatial feature map into multiple complementary views. These views are then grouped and aggregated by a block-structured, content-dependent router \cite{team2026attention,shazeer2017outrageously}, so that different subsets of spatial information contribute depending on the image. The visual target is learned jointly with the EEG encoder through EEG--vision contrastive learning and is combined with a two-stage schedule that uses MMD regularization \cite{gretton2012kernel} to reduce distributional differences among source subjects. For unseen subjects, we further introduce a post hoc \emph{representation refinement} that requires no target-subject labels or additional training. It aligns the frozen EEG and image embeddings through moment matching, CSLS-based pseudo-correspondences \cite{lample2018word}, and regularized orthogonal alignment. This step targets the residual distribution shift that remains after training, which training-time regularization cannot fully remove because target-subject data are unavailable during training. Since the refinement uses the unlabeled test embeddings of the held-out subject collectively (i.e., it is transductive), we report it separately.

Our contributions are:
\begin{itemize}
\item a \emph{structured multi-view visual target} that converts spatially distributed PE features into complementary supervision views for cross-subject EEG decoding;
\item a \emph{content-dependent block routing} strategy that adaptively aggregates these views for each image;
\item a training-free \emph{representation refinement} for unseen subjects. In ten-subject LOSO experiments on THINGS-EEG2, the structured target achieves the highest average retrieval accuracy among the methods reported in Table \ref{tab:inter_subject_full}, while the separate representation-refinement stage further raises Top-1 accuracy to
48.1\%.
\end{itemize}

\section{Method}

\subsection{Problem Setup}

We address cross-subject EEG-to-image retrieval under a leave-one-subject-out (LOSO) protocol. Let
\begin{equation}
\mathcal{D}=\{(x_i,I_i,s_i)\}_{i=1}^{N},
\end{equation}
where $x_i\in\mathbb{R}^{C\times T}$ denotes an EEG trial, $I_i$ is the corresponding image, and $s_i$ denotes the subject identity. We learn an EEG encoder $f_{\theta}$ that maps neural responses into a shared embedding space with a frozen visual encoder. During training, nine subjects are used as source subjects, while the held-out subject is completely unseen. At inference, the learned EEG encoder is applied without target-subject fine-tuning.

Rather than using a single global image descriptor for supervision, we construct a structured multi-view visual target from the spatial representation produced by the frozen visual encoder. The detailed architecture is depicted in Fig. \ref{fig:architecture}.

\subsection{Structured Multi-View Visual Target}

Let the spatial patch representation of an image be denoted by $P\in\mathbb{R}^{T\times D}$.To preserve complementary spatial information, we introduce $K$ learnable queries
$Q\in\mathbb{R}^{K\times D}$ and extract multiple visual views through cross-attention:
\begin{equation}
V=
\operatorname{MHA}
\left(
Q,\operatorname{LN}(P),\operatorname{LN}(P)
\right),
\qquad
V\in\mathbb{R}^{K\times D}.
\end{equation}
where \(P\) denotes the spatial patch features, \(Q\) the learnable queries, \(\operatorname{LN}\) Layer Normalization, and \(\operatorname{MHA}\) multi-head cross-attention.
Each query learns a distinct attention pattern over the spatial tokens, producing multiple task-specific views instead of collapsing the entire feature map into a single global vector. The extracted views are subsequently projected into the common EEG--vision embedding space:
\begin{equation}
z_k=f_k(V_k),\qquad k=1,\ldots,K.
\end{equation}

We then adapt block attention-residual routing to aggregate these visual views into a structured supervision target. This adaptation is designed specifically to construct visual supervision for cross-subject EEG representation learning, rather than to combine transformer-layer residuals.

\begin{table*}[t]
\centering
\caption{Cross-subject zero-shot EEG-to-image retrieval on THINGS-EEG2 under leave-one-subject-out evaluation. Results are reported as Top-1 / Top-5 accuracy (\%).}
\label{tab:inter_subject_full}
\scriptsize
\setlength{\tabcolsep}{3pt}
\renewcommand{\arraystretch}{1.05}
\resizebox{\textwidth}{!}{%
\begin{tabular}{lccccccccccc}
\toprule
Method
& Sub-01 & Sub-02 & Sub-03 & Sub-04 & Sub-05
& Sub-06 & Sub-07 & Sub-08 & Sub-09 & Sub-10 & Avg. \\
\midrule
NICE \cite{song2024decoding}
& 7.6/22.8 & 5.9/20.5 & 6.0/22.3 & 6.3/20.7 & 4.4/18.3
& 5.6/22.2 & 5.6/19.7 & 6.3/22.0 & 5.7/17.6 & 8.4/28.3
& 6.2/21.4 \\
ATM \cite{li2024visual}
& 10.5/26.8 & 7.1/24.8 & 11.9/33.8 & 14.7/39.4 & 7.0/23.9
& 11.1/35.8 & 16.1/43.5 & 15.0/40.3 & 4.9/22.7 & 20.5/46.5
& 11.9/33.8 \\
UBP \cite{11094846}
& 11.5/29.7 & 15.5/40.0 & 9.8/27.0 & 13.0/32.3 & 8.8/33.8
& 11.7/31.0 & 10.2/23.8 & 12.2/32.2 & 15.5/40.5 & 16.0/43.5
& 12.4/33.4 \\
NeuroBridge \cite{zhang2026neurobridge}
& 23.2/52.4 & 21.2/49.3 & 13.2/36.5 & 17.0/45.3 & 14.5/37.7
& 25.0/55.0 & 15.3/45.1 & 20.1/44.9 & 13.7/36.5 & 27.2/56.3
& 19.0/45.9 \\
Shallow Alignment \cite{du2026deep}
& 24.6/54.7 & 31.3/61.5 & 11.4/31.1 & 19.9/48.8 & 19.0/45.5
& 24.1/49.8 & 18.6/51.6 & 17.6/46.7 & 23.3/54.9 & 34.6/63.2
& 22.4/50.8 \\
% SAMGA \cite{jiang2026subject} (reported)
% & 36.3/69.8 & 42.2/71.7 & 24.7/50.7 & 34.0/66.7 & 30.0/64.3
% & 35.2/67.7 & 35.1/61.0 & 28.7/59.5 & 28.1/57.0 & 49.6/79.4
% & 34.4/64.8 \\
SAMGA$^{*}$ \cite{jiang2026subject}
& 36.0/58.5 & 37.5/68.5 & 21.0/43.0 & 29.0/56.0 & 21.0/51.0
& 32.0/62.0 & 25.5/55.5 & 26.0/50.0 & 24.0/51.0 & 43.5/72.5
& 29.6/56.8 \\
\midrule
\textbf{Ours}
& 41.5/72.0 & 38.5/73.0 & 24.5/51.0 & 39.0/71.5 & 32.5/61.0
& 35.5/64.5 & 33.5/69.0 & 32.5/58.0 & 27.5/55.5 & 48.0/80.5
& \textbf{35.3/65.6} \\
\textbf{Ours$^{\dagger}$}
& \textbf{56.0/84.0} & \textbf{58.0/84.0} & \textbf{41.5/72.5}
& \textbf{44.0/75.5} & \textbf{44.5/74.5}
& \textbf{49.5/76.0} & \textbf{48.0/78.0} & \textbf{37.0/63.0}
& \textbf{43.5/76.0} & \textbf{59.0/87.5}
& \textbf{48.1/77.1} \\
\bottomrule
\end{tabular}%
}
\vspace{-0.2cm}
\begin{flushleft}
\scriptsize $^{*}$Reproduced using the official implementation under the same experimental settings as the proposed method.
$^{\dagger}$Transductive: uses unlabeled target-subject embeddings without labels or encoder updates for representation refinement.
\end{flushleft}
\vspace{-0.5cm}
\end{table*}

\subsection{Block Attention Residual View Routing}

The learned visual views are hierarchically aggregated through block attention-residual routing to form a structured visual supervision target for cross-subject EEG--vision learning \cite{team2026attention}. The $K$ views are divided into $B$ equal-sized blocks. Within each block, a learnable query $q_v$ computes content-dependent attention weights:
\begin{equation}
a_{b,k} =
\operatorname{softmax}_{k\in\mathcal{B}_b}
\left(
\frac{q_v^\top \operatorname{LN}(z_k)}{\tau}
\right),
\end{equation}
where $\mathcal{B}_b$ denotes the set of views in the $b$-th block. The block representation is
\begin{equation}
h_b =
\sum_{k\in\mathcal{B}_b}
a_{b,k}z_k.
\end{equation}

A second learnable query $q_B$ assigns content-dependent weights
to the block representations:
\begin{equation}
w_b =
\operatorname{softmax}_b
\left(
\frac{q_B^\top \operatorname{LN}(h_b)}{\tau}
\right).
\end{equation}

The final view weight is
\begin{equation}
\omega_{b,k}=w_ba_{b,k}.
\end{equation}

The structured visual target is then obtained as
\begin{equation}
z_I =
g\left(
\sum_{b=1}^{B}
\sum_{k\in\mathcal{B}_b}
\omega_{b,k}z_k
\right).
\end{equation}

where $g(\cdot)$ denotes the final projection into the shared EEG--vision embedding space.

In our implementation, $K=12$ visual views are organized into $B=4$ blocks. This two-level aggregation selectively integrates complementary visual views while avoiding a single flat softmax over all views. The key contribution is the adaptation of block attention-residual routing to structured visual target construction for cross-subject EEG--vision representation learning.

\subsection{Training Objective}
\label{method}
We optimize the EEG and visual projection components using a temperature-scaled contrastive EEG--image objective. To improve cross-subject consistency, training uses a two-stage optimization schedule. In Stage~I, the contrastive objective is jointly optimized with an MMD-based regularization term that reduces distributional differences among source subjects. The MMD contribution is progressively reduced during this stage. In Stage~II, the shared projection layer is frozen and the optimization focuses on EEG--vision contrastive alignment with a reduced learning rate.

This training strategy encourages subject-invariant EEG representations while preserving the semantic alignment required for image retrieval. For the representation-refinement setting, the trained encoders are kept frozen and refinement is performed post hoc on the held-out subject using only the unordered EEG and image embeddings, without target-subject labels or additional model training. The procedure first performs per-dimension moment matching, followed by pseudo-correspondence estimation using CSLS and mutual nearest-neighbor landmarks \cite{lample2018word}. We use 64 landmarks and confidence-based weighting for the estimated correspondences, followed by regularized orthogonal alignment with a regularization coefficient $\rho=0.1$. The resulting aligned
representations are finally evaluated using CSLS-based retrieval. This post hoc procedure is referred to as \emph{representation refinement} and is evaluated separately from the trained model's direct output.

\section{Experiments}

\subsection{Dataset Details}

We evaluate on the THINGS-EEG2 dataset \cite{gifford2022large}, which contains EEG recordings from 10 subjects collected using a rapid serial visual presentation (RSVP) paradigm. The training set comprises 1,654 object concepts, with 10 images per concept and 4 repetitions per image, resulting in 16,540 images and 66,160 EEG trials per subject. The test set contains 200 previously unseen concepts, with one image per concept repeated 80 times, forming a shared 200-image retrieval gallery. EEG recordings consist of 63 channels and are downsampled from 1,000 Hz to 250 Hz for modeling. We follow a strict ten-fold leave-one-subject-out (LOSO) protocol, where nine subjects are used for training and the remaining subject is held out entirely for testing. We report 200-way zero-shot Top-1 and Top-5 retrieval accuracy.

\subsection{Evaluation Details}
For each EEG trial, the learned EEG encoder produces an embedding
\begin{equation}
z_E=f_s\left(f_p\left(f_{\theta}(x_t)\right)\right),
\end{equation}
where $f_{\theta}$ denotes the EEG encoder and $f_p,f_s$ denote the projection modules. Each candidate image is represented by its corresponding structured visual target, forming the retrieval gallery
\begin{equation}
\mathcal{G}
=
\left\{
z_I^{(1)},z_I^{(2)},\ldots,z_I^{(M)}
\right\}.
\end{equation}
Retrieval is performed by ranking the gallery according to cosine similarity,
\begin{equation}
S(z_E,z_I)
=
\frac{z_E^\top z_I}
{\|z_E\|_2\|z_I\|_2}.
\end{equation}
The highest-scoring image determines Top-1 retrieval, while the five highest-scoring images determine Top-5 retrieval.

\subsection{Implementation Details}

All experiments are conducted on a Linux-based system equipped with three NVIDIA RTX A6000 GPUs, each providing 48 GB of GPU memory. The EEG signal is encoded using a TSConv-based neural encoder followed by a linear projection into a 512-dimensional shared EEG--vision embedding space \cite{song2024decoding}. The encoder operates on a $0$--$250$ ms time window, with smoothing-based EEG augmentation enabled during training, while the visual encoder (Perception Encoder PE-Core-G14
\cite{bolya2026perception}) is kept frozen. We extract $K=12$ visual views from the spatial visual representation and organize them into $B=4$ blocks for hierarchical block attention-residual routing. The router uses a temperature of $0.7$ and a subject-bias dropout rate of $0.3$, with EEG-conditioned routing disabled.

Training uses a batch size of $1024$ and an initial learning rate of $1\times10^{-4}$ for up to 50 epochs. The contrastive temperature is fixed at $0.07$, with $\alpha=\beta=1.0$ for the alignment objectives. Image features are $\ell_2$-normalized before alignment. Following the two-stage optimization described in Section~\ref{method}, Stage I runs for 20 epochs and jointly optimizes the contrastive and distribution-alignment objectives, with the distribution-alignment weight reduced from $0.9$ to $0.5$. In Stage II, the shared projection component is frozen and contrastive training continues with a reduced learning rate of $5\times10^{-5}$. Early stopping is applied with respect to the validation set with a patience of 10 epochs. The EEG feature dimension before projection is 1024, and the intermediate visual feature dimension is 1024.

For cross-subject evaluation, we follow a leave-one-subject-out protocol on THINGS-EEG2, using nine subjects for source training and the remaining subject for evaluation. All experiments use a fixed random seed of 2025. The frozen visual representation uses the specified spatial feature representation with visual layers 20, 24, 28, 32, and 36 as the source features for multi-view target construction.

\subsection{Results and Discussion}

\begin{figure}[t]
    \centering
    \includegraphics[width=\columnwidth]{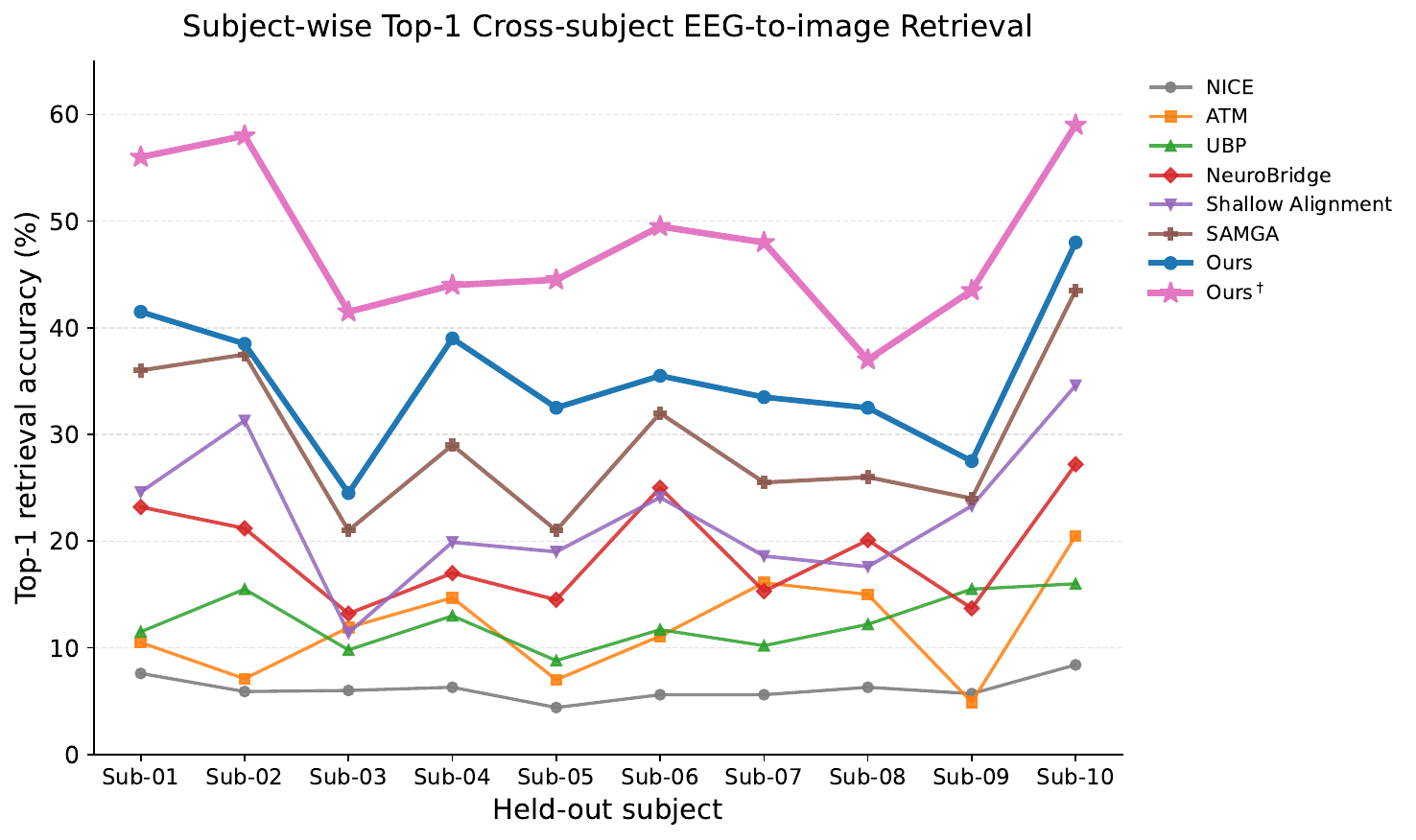}
    \caption{Subject-wise Top-1 retrieval accuracy on THINGS-EEG2 under
    leave-one-subject-out evaluation. Results are reported for each of the
    ten held-out subjects across all compared methods.}
    \label{fig:subject_wise_top1}
    \vspace{-0.2cm}
\end{figure}

% Figure~\ref{fig:subject_wise_top1} shows the subject-wise Top-1 retrieval
% performance across the ten held-out subjects.

\begin{figure}[t]
    \centering
    \includegraphics[width=\columnwidth]{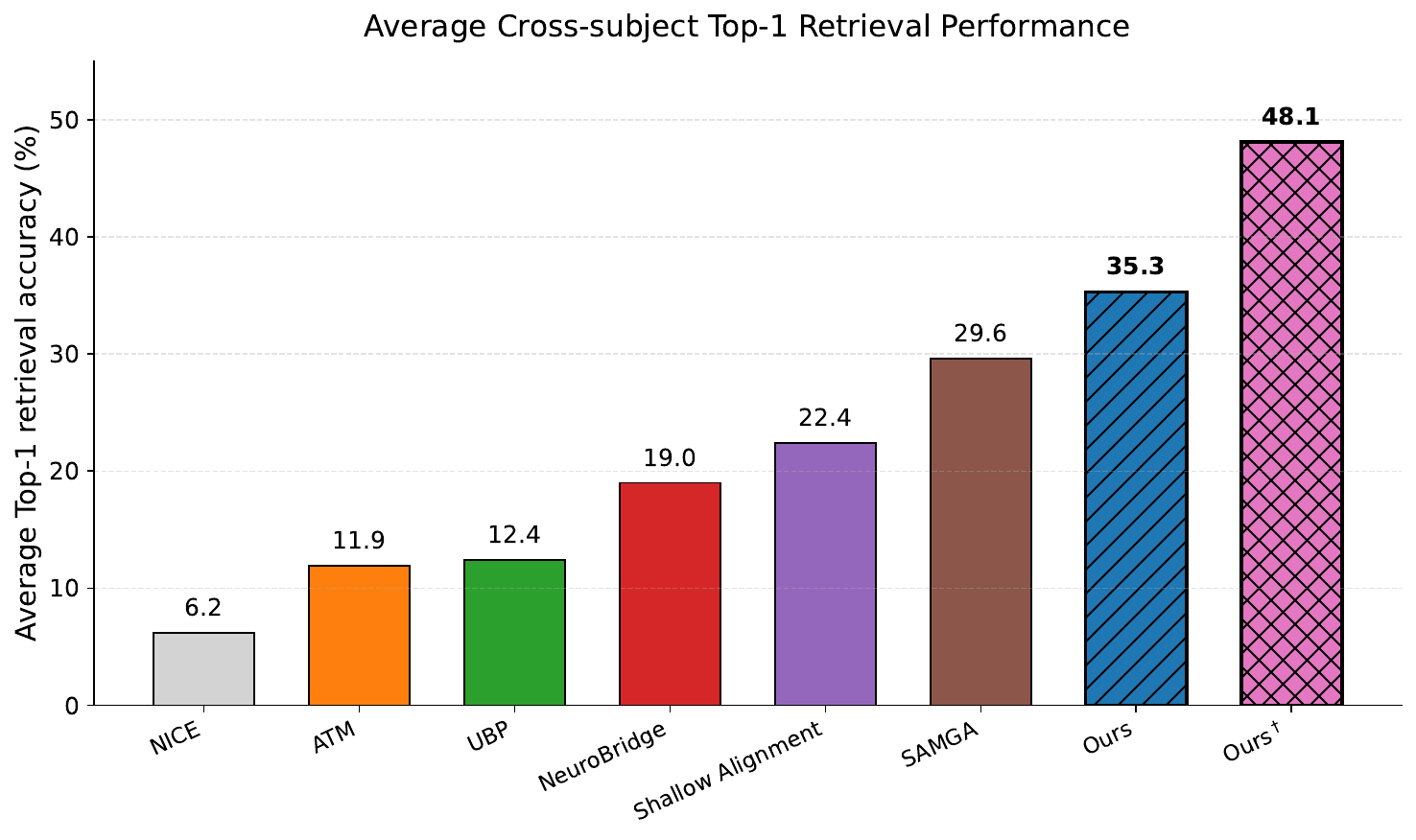}
    \caption{Average cross-subject Top-1 retrieval accuracy on THINGS-EEG2
    under leave-one-subject-out evaluation. Results are averaged across the
    ten held-out subjects for all compared methods.}
    \label{fig:average_top1}
    \vspace{-0.3cm}
\end{figure}

\begin{figure}[t]
    \centering
    \includegraphics[width=\columnwidth]{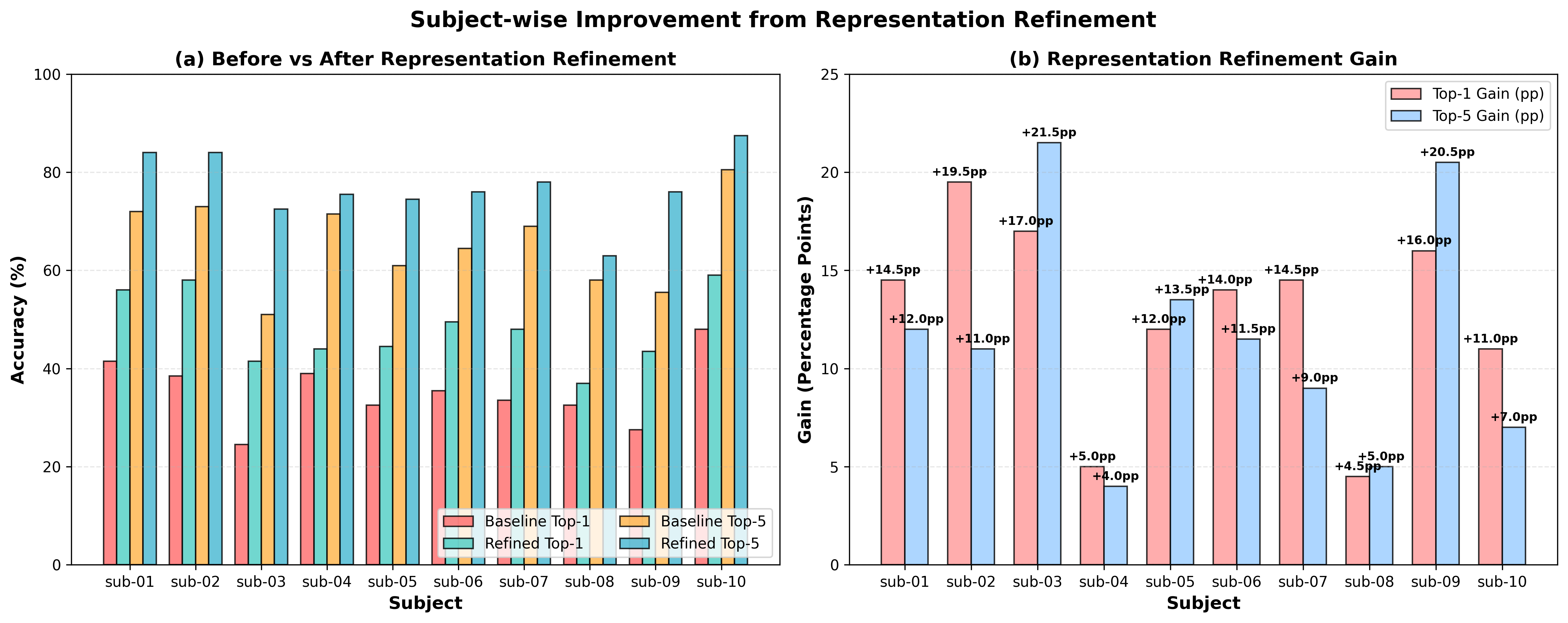}
    \caption{Subject-wise effect of representation refinement on cross-subject EEG-to-image retrieval. 
    (a) Top-1 and Top-5 retrieval accuracy before and after refinement for each held-out subject.
    (b) Subject-wise Top-1 and Top-5 accuracy gains, reported in percentage points (pp).}
    \label{fig:refinement_subjectwise}
    \vspace{-0.3cm}
\end{figure}

\textbf{Comparison with prior methods.} Table~\ref{tab:inter_subject_full} reports LOSO retrieval accuracy for all ten held-out subjects. Without any target-subject adaptation, the proposed structured visual target achieves 35.3\% Top-1 and 65.6\% Top-5 accuracy on average, the highest among the compared methods. For a controlled comparison with the strongest baseline, we reproduced SAMGA~\cite{jiang2026subject} using its official implementation under our experimental settings (SAMGA$^{*}$). Under identical settings, our model improves the average accuracy by 5.7 points in Top-1 and 8.8 points in Top-5, and outperforms SAMGA$^{*}$ on all ten held-out subjects for both metrics. Fig.~\ref{fig:subject_wise_top1} shows this comparison per subject. All methods follow a similar difficulty profile: Sub-10 is the easiest held-out subject for every method, and Sub-03 is among the hardest for the stronger methods, reflecting subject-dependent signal quality. Nevertheless, our model stays above SAMGA$^{*}$ for every subject, and the refined model is the best for every subject. As summarized in Fig.~\ref{fig:average_top1}, the average Top-1 accuracy rises from 29.6\% for SAMGA$^{*}$ to 35.3\% with the structured target and to 48.1\% with refinement. This suggests that supervising EEG with multiple learned visual views, rather than a single global image embedding, yields representations that transfer better to unseen subjects.

% \textbf{Comparison with prior methods.} Table~\ref{tab:inter_subject_full} reports LOSO retrieval accuracy for all ten held-out subjects. Without any target-subject adaptation, the proposed structured visual target achieves 35.3\% Top-1 and 65.6\% Top-5 accuracy on average, the highest among the compared methods and above the results reported for SAMGA~\cite{jiang2026subject} (34.4\%/64.8\%). To control for differences in experimental settings, we also reproduced SAMGA using its official implementation under our settings (SAMGA$^{*}$). Under identical settings, our model improves the average accuracy by 5.7 points in Top-1 and 8.8 points in Top-5, and outperforms SAMGA$^{*}$ on all ten held-out subjects for both metrics. Fig.~\ref{fig:subject_wise_top1} shows this comparison per subject. All methods follow a similar difficulty profile: Sub-10 is the easiest held-out subject for every method, and Sub-03 is among the hardest for the stronger methods, reflecting subject-dependent signal quality. Nevertheless, our model stays above SAMGA$^{*}$ for every subject, and the refined model is the best for every subject. On average, Top-1 accuracy rises from 29.6\% for SAMGA$^{*}$ to 35.3\% with the structured target and to 48.1\% with refinement. This suggests that supervising EEG with multiple learned visual views, rather than a single global image embedding, yields representations that transfer better to unseen subjects.

\textbf{Representation refinement.} Applying the label-free refinement to the frozen embeddings raises the average accuracy from 35.3\% to 48.1\% Top-1 and from 65.6\% to 77.1\% Top-5. Because the refinement uses the unlabeled test embeddings of the held-out subject collectively, we report it separately from the direct model output. As shown in Fig.~\ref{fig:refinement_subjectwise}, refinement improves both Top-1 and Top-5 accuracy for every held-out subject, with Top-1 gains ranging from 4.5 percentage points (pp) for Subject~8 to 19.5~pp for Subject~2 (mean 12.8~pp). The gain is only weakly related to a subject's accuracy before refinement (Pearson $r=-0.23$ across subjects), suggesting that the benefit depends on the subject-specific structure of the embeddings rather than on how poorly the subject is initially decoded.

\textbf{Discussion.} The two components act at different stages. The structured target shapes the EEG representation during source training, whereas refinement aligns the resulting embeddings to an unseen subject at deployment without labels or encoder updates. The large refinement gains indicate that, for new subjects, the source-trained embeddings retain stimulus-related structure that is misaligned rather than lost. A limitation is that refinement currently uses all test trials of the held-out subject; its behavior with smaller target batches remains to be characterized.

\section{CONCLUSION}

We presented a structured visual target learning framework for cross-subject
EEG-to-image retrieval. The proposed approach preserves spatial information
from the frozen Perception Encoder, converts it into multiple learned visual
views, and adaptively aggregates them through block-structured,
content-dependent routing. MMD-based regularization further promotes
cross-subject consistency during training. We also introduced a training-free
representation refinement stage that aligns frozen EEG and visual embeddings
for an unseen subject without target-subject labels or encoder updates.
Future work will explore refinement with limited target-subject data and
alternative visual backbones.

\bibliographystyle{IEEEbib}

\bibliography{strings,refs}

\end{document}